%% file: main.tex
\documentclass[runningheads]{llncs}

\usepackage[T1]{fontenc}
\def\doi#1{\href{https://doi.org/\detokenize{#1}}{\url{https://doi.org/\detokenize{#1}}}}
\usepackage{graphicx}
\usepackage{multirow}
\usepackage{amssymb}
\usepackage{pifont}
\usepackage{listings}
\usepackage{xcolor}
\usepackage{tabularx}
\usepackage{ctable}
\usepackage{bbm}
\usepackage{amsmath}
\usepackage{subfig}

\usepackage[symbol]{footmisc}

\newcommand{\Bparagraph}[1]{\noindent\textbf{#1}}

\begin{document}
\title{Did You Get What You Paid For? Rethinking Annotation Cost of Deep Learning Based Computer Aided Detection in Chest Radiographs}
\author{
Tae Soo Kim\thanks{\textit{Authors equally contributed.}},
Geonwoon Jang$^*$,
Sanghyup Lee,
Thijs Kooi
}
%index{Kim, Tae Soo}
%index{Jang, Geonwoon}
%index{Lee, Sanghyup}
%index{Kooi, Thijs}

%
\authorrunning{Kim et al.}
\institute{Lunit Inc.}
\maketitle              % typeset the header of the contribution
\begin{abstract}
\input{abstract}
\end{abstract}
\section{Introduction}
\input{introduction}

\section{Methods}
\input{methods}

\section{Experiments}

\input{experiments}

\section{Discussion and Conclusions}
\input{conclusion}

\bibliographystyle{splncs04}
\bibliography{refs}
\end{document}

\title{Did You Get What You Paid For? Rethinking Annotation Cost of Deep Learning Based Computer Aided Detection in Chest Radiographs}
\author{Supplementary material for Paper ID 682}
\institute{Lunit Inc.}

\maketitle              % typeset the header of the contribution
%
%\section{Additional Information on Data}
%\vspace{-0.5cm}

\begin{table}[h]
\centering
\caption{Additional details regarding the internal dataset used in this study. CR: Chest radiograph. ($^*$Omitted during blind-review process.)}
\begin{tabular}{l|l}
\toprule
  &  Atelectasis, Calcification, Cardiomegaly, Consolidation, Fibrosis,\\ 
 Chest abnormalities &  Mediastinal widening, Nodule, Pleural Effusion,\\
 & Pneumoperitoneum, Pneumothorax. \\\hline
  & The data was collected from XXX$^*$ Hospital between \\
 Data collection site & March 2004 and December 2017. The dataset additionally \\
 & includes raw CRs collected from public sources that contain \\
 & data from Country 1, Country 2 and Country 3$^*$.\\ \hline
  & All CRs were reviewed by at least \\
Labeling group & one of 20 board-certified radiologists (labeling-group) with 7-14 \\
 &years of experience in reading CRs. The CRs from external sources \\
 & were re-annotated by the same labeling group.
\end{tabular}
\end{table}
\vspace{-0.8cm}

\begin{table}[h]
\centering
\caption{Additional details regarding various experimental settings.}
\label{tab:supp_setting}
\begin{tabular}{c | c | c| c | c| c }
\toprule
 Epochs & Learning Rate & Batch Size & Weight Decay& Optimizer & Framework \\ \midrule
 40 & 0.04 for 0-30 epochs & 256 & 0.0001 & SGD  & PyTorch\\
    & 0.004 for 30-40 epochs &&& with momentum 0.9&
\end{tabular}
\end{table}
% The chest radiographs (CRs) with normal or abnormal findings were initially collected and classified in terms of whether they had each of the 10 abnormalities: Atelectasis, Calcification, Cardiomegaly, Consolidation, Fibrosis, Mediastinal widening, Nodule, Pleural Effusion, Pneumoperitoneum, Pneumothorax. The data was collected from XXX\footnote{Omitted for double-blind review.} Hospital between March 2004 and December 2017. The dataset additionally includes raw CRs collected from public sources that contain data from Country 1, Country 2 and Country 3\footnote{Omiited for double-blind review.}. For data curation, all CRs were reviewed by at least one of 20 board-certified radiologists (labeling-group) with 7-14 years of experience in reading CRs. The CRs from external sources were re-annotated by the same labeling group. The annotation process was performed in two steps: image-level classification and pixel-level annotation. 

%\vspace{-1.0cm}
%\section{Additional Experimental Results with Different Backbones}
%\vspace{-0.5cm}

\begin{table*}[h]
\centering
\setlength{\tabcolsep}{2.5pt}
\caption{Comparison of different ResNet backbones and their classification/segmentation performance differences. All results are trained with both image-level and pixel-level labels.
We train the models under the same settings as Table~\ref{tab:supp_setting}, only with smaller batch size of 80 for ResNet101 due to our resource limits.
}
\label{tab:supp_backbones}
\begin{tabular}{c|ccccc}
 & \multicolumn{5}{c}{Dataset Size} \\
Classification & 6K (5\%) & 12K (10\%) & 30K (25\%) & 60K (50\%) & 121K \\
\toprule
Res18 & 87.3 $\pm$ 0.9 & 91.7 $\pm$ 0.3 & 94.3 $\pm$ 0.1 & 95.00 $\pm$ 0.1 & 95.5 $\pm$ 0.08 \\
Res34 & 81.2 $\pm$ 1.2 & 88.5 $\pm$ 0.7 & 94.3 $\pm$ 0.1  & 94.9 $\pm$ 0.3 & 95.8 $\pm$ 0.01 \\
Res101 & 91.0 $\pm$ 0.6 & 94.0 $\pm$ 0.5 & 95.4 $\pm$ 0.1 & 95.5 $\pm$ 0.3 & 95.8 $\pm$ 0.03 \\
\midrule
Segmentation & & & & & \\
\toprule
Res18 & .307 $\pm$ .005 & .374 $\pm$ .01 & .443 $\pm$ .004 & .464 $\pm$ .009 & .477 $\pm$ .002 \\
Res34 & .327 $\pm$ .03 & .422 $\pm$ .005 & .469 $\pm$ .009 & .486 $\pm$ .009 & .500 $\pm$ .007 \\
Res101 & .363 $\pm$ .02 & .415 $\pm$ .02 & .465 $\pm$ .005 & .469 $\pm$ .007 & .464 $\pm$ .002 \\
\bottomrule
% Image-level & 6K (5\%) & 12K (10\%) & 30K (25\%) & 60K (50\%) & 121K \\
% \midrule
% Res18 & 78.5 $\pm$ 4.3 & 85.3 $\pm$ 3.1 & 93.1 $\pm$ 0.2 & 94.5 $\pm$ 0.1 & 95.4 $\pm$ 0.04 \\
% % Res34 & $\pm$ & $\pm$ & $\pm$ & $\pm$ & $\pm$ \\
% Res34 & 84.0 $\pm$ 3.2 & 90.4 $\pm$ 0.1  & 92.4 $\pm$ 0.2 & 92.9 $\pm$ 0.1  & 93.7 $\pm$ 0.7 \\
% Res101 & 87.4 & 92.1 & 94.9 & 95.4 & 95.5 \\
% \midrule
\end{tabular}
\end{table*}

\begin{figure}[h]
\centering
    \renewcommand{\wp}{0.17 \linewidth}
    \centering
    % \subfloat[Contour 121K]{\includegraphics[width=\wp]{figures/vis_comparison/contour100/merged_col.png}}
    \subfloat[]{\includegraphics[width=\wp]{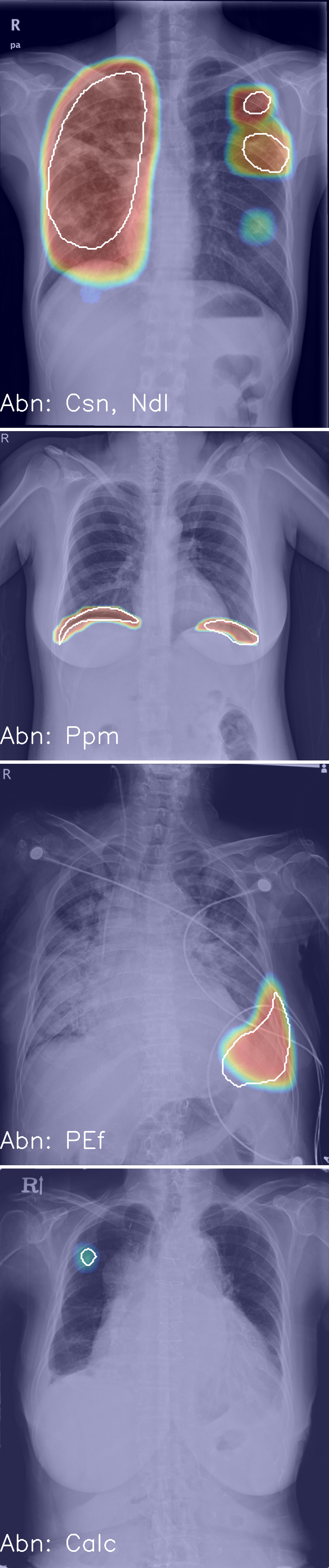}}
    \hspace{0.005 \linewidth}
    % \subfloat[Bbox 121K]{\includegraphics[width=\wp]{figures/vis_comparison/bbox100/merged_col.png}}
    \subfloat[]{\includegraphics[width=\wp]{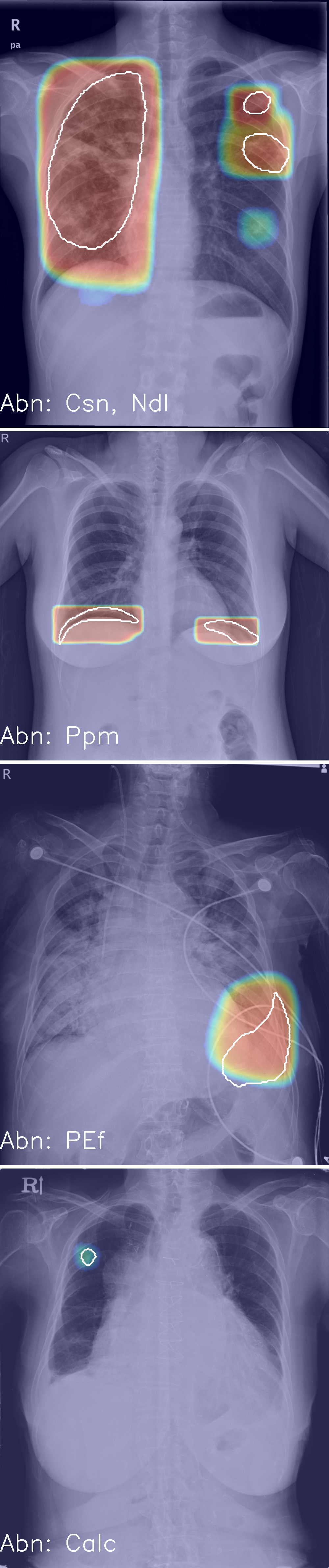}}
    \hspace{0.005 \linewidth}
    % \subfloat[Contour 6K + Bbox 115K]{\includegraphics[width=\wp]{figures/vis_comparison/contour05-bbox95/merged_col.png}}
    \subfloat[]{\includegraphics[width=\wp]{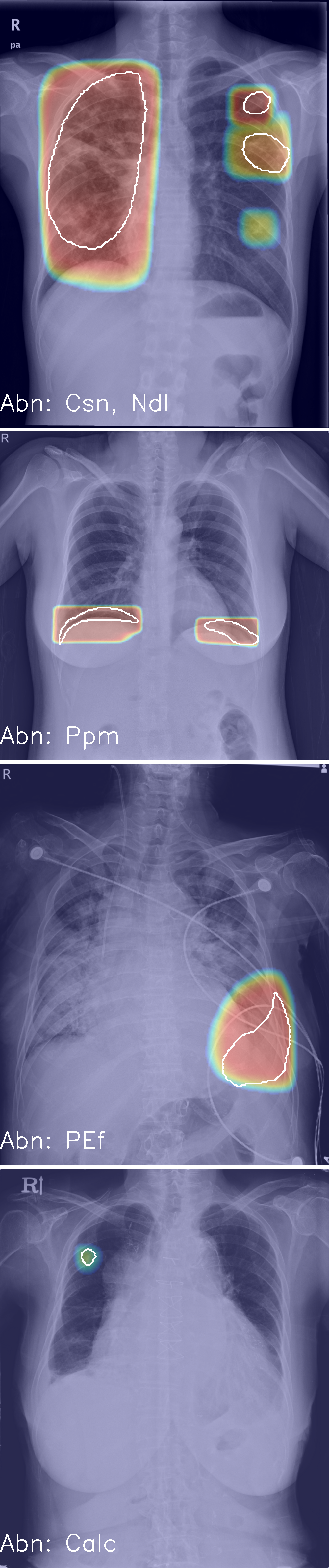}}
    \hspace{0.005 \linewidth}
    % \subfloat[Image-level 6K + Bbox 115K]{\includegraphics[width=\wp]{figures/vis_comparison/contour05-cls95/merged_col.png}}
    \subfloat[]{\includegraphics[width=\wp]{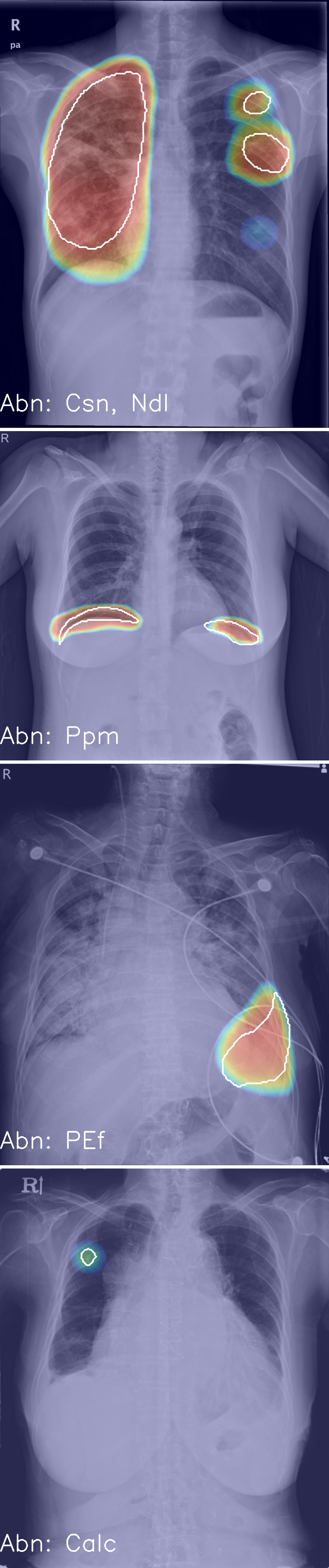}}
    \caption{Comparison of segmentation models trained with 
    (a) 121K gold-standard contours (GSC), 
    (b) 121K bounding-box pixel-level labels, 
    (c) 6K GSC and 115K bounding-box pixel-level labels, and
    (d) 6K GSC and 115K image-level labels.
    The existing abnormalities are tagged in abbreviations (PEf, Ndl, Csn etc) of Table 1 in the main manuscript.
    Best viewed in color.
    }
    \label{fig:supp_seg_vis_comp}
\end{figure}

%\bibliographystyle{splncs04}
%\bibliography{refs}

%% file: abstract.tex
% Broader impact
%Deep neural networks have enabled powerful computer aided detection (CAD) systems for finding abnormalities in chest radiographs. 
% Set up / the bridge
As deep networks require large amounts of accurately labeled training data, a strategy to collect sufficiently large and accurate annotations is as important as innovations in recognition methods. This is especially true for building Computer Aided Detection (CAD) systems for chest X-rays where domain expertise of radiologists is required to annotate the presence and location of abnormalities on X-ray images.
% The problem
However, there lacks concrete evidence that provides guidance on how much resource to allocate for data annotation such that the resulting CAD system reaches desired performance. Without this knowledge, practitioners often fall back to the strategy of collecting as much detail as possible on as much data as possible which is cost inefficient.
% Contribution
In this work, we investigate how the cost of data annotation ultimately impacts the CAD model performance on classification and segmentation of chest abnormalities in frontal-view X-ray images.
% Method & Results
We define the cost of annotation with respect to the following three dimensions: quantity, quality and granularity of labels. Throughout this study, we isolate the impact of each dimension on the resulting CAD model performance on detecting 10 chest abnormalities in X-rays.
%We relate cost to the amount of labor required by a radiologist during the annotation job and define four commonly used annotation methods with varying costs: image-level annotations derived from text-reports and image-level, bounding-box, contour level annotations from radiologists.
On a large scale training data with over 120K X-ray images with gold-standard annotations, we find that cost-efficient annotations provide great value when collected in large amounts and lead to competitive performance when compared to models trained with only gold-standard annotations. We also find that combining large amounts of cost efficient annotations with only small amounts of expensive labels leads to competitive CAD models at a much lower cost.

% CAD systems are powerful and are driven by deep learning models.
% Deep nets require accurately labeled data in large amounts.
% But no one really knows how accurate the data should be and how much data is actually required for certain level of performance -- usually, 
% In this paper, we investigate how much 

%% file: introduction.tex
% Broader impact
Deep neural network driven Computer Aided Detection (CAD) systems for chest X-ray images have become powerful aids to radiologists and have been shown to surpass human performance \cite{Jin2022}. 
% Set up / the bridge
One of the biggest driving forces behind the innovation is the collection of large scale chest X-ray datasets with accurate annotations provided by expert radiologists \cite{cheXpert,wang_cxr14,CALLI2021}. 
% The problem
However, curation of accurately annotated large scale datasets for CAD systems in X-rays is challenging because the expertise of radiologists is required to annotate abnormalities.
% the existence of chest abnormalities in X-rays.

% How people have addressed it and shortcomings.
Without access to radiologists, an alternative is to generate labels by parsing radiology reports and training a language model to predict the existence of findings from associated text reports~\cite{cheXpert,CheXbert_emnlp2020,wang_cxr14}. Although this can minimize the cost of annotation, the performance of the resulting CAD system inevitably hinges on the accuracy of the generated labels \cite{Rayner2019}, as we also later verify in this paper. There has been attempts to leverage unlabeled data \cite{Li2020SelfLoopUA} and use mixed set of supervision \cite{KARIMI2020} which can also contribute towards saving annotation cost. However, it is still unclear how different dimensions of annotation cost quantitatively impact the resulting CAD system performance. 
%Moreover, the language model itself requires an initial set of gold standard annotations provided by radiologists \cite{CheXbert_emnlp2020}. 
%Even when practitioners have access to radiologists, collecting a large set annotations from radiologists is far more costly than collecting equivalent ground truth on natural images. 
%Annotation cost undeniably affects the entire development process of X-ray CAD models yet it is still unclear how the cost of annotation quantitatively impacts the resulting CAD system performance. 

% Contributions
In this work, we investigate how annotation cost impacts the performance of CAD systems for classifying and localizing ten chest abnormalities in frontal view (PA) chest X-ray. We define the total cost of annotation (i.e., the effort it takes to annotate a scan) as a function of three factors: the \textit{quantity}, \textit{quality} and \textit{granularity} of annotations. We highlight some of the key motivations and insights provided by this study with respect to each dimension of annotation cost.

\begin{enumerate}
    \item \textbf{Granularity: } A radiologist can curate a given chest radiograph at an image-level and pixel-level. We further decompose the pixel-level granularity by the amount of effort and thus time required by a radiologist to annotate a given image. For example, a radiologist drawing accurate contours around chest abnormalities will require more time to annotate the same amount of chest radiographs than the radiologist defining a rough bounding box around the lesion. The quickest and the lowest level of granularity is to collect only image-level annotations (whether the lesion exists or not). We quantify the trade-off between CAD model performance and the added cost of collecting granular labels.
    \item \textbf{Quality:} A more cost efficient alternative to the radiologist-provided gold standard is to get image-level annotations automatically by extracting labels using natural language processing (NLP) algorithms \cite{cheXpert,CheXbert_emnlp2020}. The caveat of such algorithms is that they are vulnerable to multiple factors \cite{olatunji_2019} that lead to inaccurate labeling of X-ray images \cite{cheXpert,KARIMI2020,Rayner2019}. How do annotation quality and the amount of label noise impact the CAD model performance?
    \item \textbf{Quantity: } Lastly, the cost of annotation is determined by the number of scans that are annotated. Although several papers investigate how the number of annotated samples affect the model's performance under label noise \cite{KARIMI2020} and limited/mixed supervision \cite{Peng2021MedicalIS}, there has not been a sufficiently large enough set of X-ray images with gold standard annotations to enable analysis at the scale presented in this paper. 
    Our study includes more than four times as many annotations from radiologists than the currently available largest dataset with annotations from experts \cite{bustos2020padchest}. Furthermore, all images used in this study contain pixel-level labeling of radiologist's opinion on the location of the findings.
    %The scale of the data enables us to design experiments to test whether combining only a small set of high-cost annotations with large amounts of cost-efficient labels is a more economic strategy for building CAD systems for detecting chest abnormalities from X-rays. 
\end{enumerate}
To the best of our knowledge, this is the first study performed at this scale with over 120K gold standard annotations where we can quantitatively measure how CAD model performance is affected by the quantity, quality, and granularity of annotations.

%% file: methods.tex
% \subsection{Classification and Segmentation Methods}

% Our method consists of both pathways getting img-level cls loss and pixel-level loc loss.

% Task-wise 
To investigate the effect of different types of annotation on multi-label classification and segmentation performance, we make use of a general, flexible, and model agnostic framework. 
We first describe our simple neural network approach that we use for both tasks in Section \ref{sec:architecture}.
% and \ref{sec:learning_objective}.
Then, we describe in detail the different annotation types used in this study in Section \ref{sec:annotation_types}. 

\subsection{Common Architecture}
\label{sec:architecture}
We use the same architecture for all experiments for both classification and segmentation tasks in this study. The architecture is similar to FCN \cite{fcn_2015} with a common Residual Network (ResNet-34) backbone \cite{He2016DeepRL} but outputs a smaller spatial resolution segmentation results.

\begin{figure}[t]
\centering
     \includegraphics[width=0.85\textwidth]{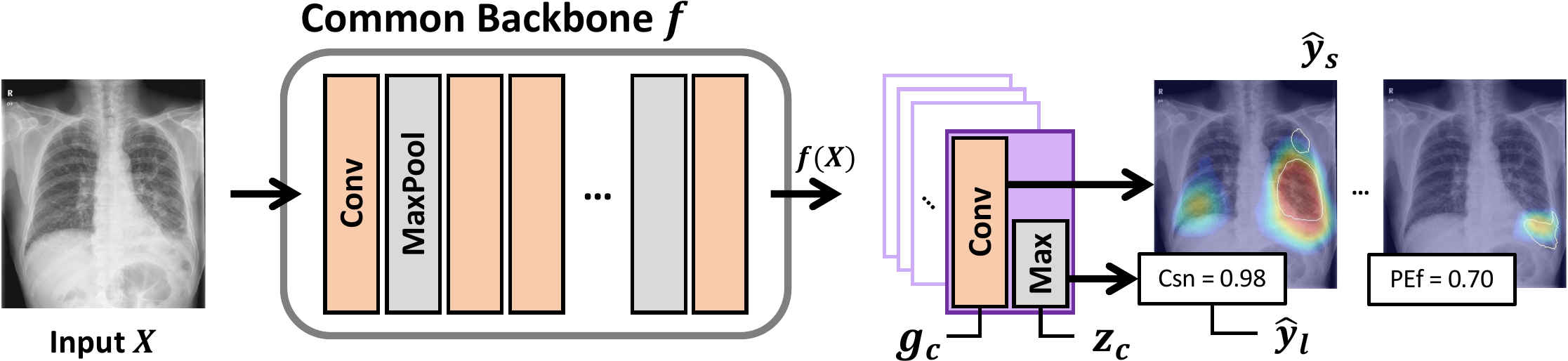}
      \caption{Overview of the model architecture. We use common backbone architectures for $f$ and append per-class classification and segmentation heads.}
      \label{fig:model}
\end{figure}

Let $X \in \mathbb{R}^{1 \times H \times W}$ be a CXR image input to our model.
The model is a neural network with a feature extractor $f : \mathbb{R}^{1 \times H \times W} \mapsto \mathbb{R}^{d \times h \times w} $ which maps an input CXR image to a feature space defined by $d$ channels and down-sampled $h,w$ spatial dimensions.
Given the feature $f(X)$, the segmentation head $g: \mathbb{R}^{d \times h \times w} \mapsto \mathbb{R}^{C \times h \times w}$ produces a pixel-wise prediction of whether a pixel belongs to one of $C$ classes. The classification head $z: \mathbb{R}^{C \times h \times w} \mapsto \mathbb{R}^{C}$ maps $g(f(X))$ to a multi-label classification of existence of chest abnormalities.
Figure \ref{fig:model} depicts the architecture used to evaluate the cost of annotation for both tasks in this study.

% \textit{Learning Objective:} 
% \subsection{Learning Objective}
% \label{sec:learning_objective}
\Bparagraph{Learning Objective:}
For the $n$-th training sample, we define the model's final loss function $\mathcal{L}$ as a sum of the classification loss $\mathcal{L}_{cls}(y_{l}^n,\hat{y}_{l}^n)$ and the segmentation loss $\mathcal{L}_{seg}(y_{s}^n,\hat{y}_{s}^n)$, where $y_{l}^n$, $\hat{y}_{l}^n$ are the terms of ground truth label and the prediction for classification and $y_{s}^n, \hat{y}_{s}^n$ are the segmentation terms. The classification loss $\mathcal{L}_{cls}(y_{l}^n,\hat{y}_{l}^n)$ is defined as a simple binary cross-entropy loss commonly used for multi-label classification problems.
% \begin{equation}
%     \mathcal{L}_{cls}(y^n, \hat{y}^n) = \frac{1}{C} \sum_{c=1}^{C} 
%     \left[ - y^n_c \log (\hat{y}^n_c) - (1-y^n_c )\log (1-\hat{y}^n_c) \right]
% \end{equation}
% where $\hat{y}^n = z(g(f(X^n))$ is the multi-label classification output of the model, $y^n_c$ is the image-level ground truth label for the $c$-th class of the $n$-th training sample and $\hat{y}^n_c$ is the corresponding model prediction. 

When pixel-wise ground truth for the $n$-th sample $y_{s}^n(h,w)$ is available for class $c$ such that $y_{s}^n(h,w) = \{0,1\}$, the model can receive additional supervision using the segmentation prediction head. We define the total segmentation loss as:
\begin{equation}
    \mathcal{L}_{seg}(y_{s}^n, \hat{y}_{s}^n) = \mathcal{L}_{pbce} + \mathcal{L}_{dice}
\end{equation}
% \begin{align}
%     \mathcal{L}_{seg}(y^n, \hat{y}^n) = \frac{1}{CHW}\sum_{c,h,w}
%     \text{BCE} ( y^n_c(h,w), \hat{y}^n_c(h,w) ) \\
%     \text{BCE}(\cdot) = - y^n_c(h,w) \log (\hat{y}^n_c(h,w)) - (1-y^n_c )\log (1-\hat{y}^n_c(h,w))
% \end{align}
% \begin{equation}
% \begin{split}   % \begin{multline}
%     \mathcal{L}_{seg}(y^n, \hat{y}^n) = \frac{1}{CHW}\sum_{c,h,w}
%     [ - y^n_c(h,w) \log (\hat{y}^n_c(h,w)) \\
%     - (1-y^n_c(h,w) )\log (1-\hat{y}^n_c(h,w)) ]
% \end{split}     % \end{multline}
% \end{equation}
% \begin{equation}
%     \mathcal{L}_{seg}(y^n, \hat{y}^n) = 
%     \frac{1}{CHW}\sum_{c=1}^{C} \sum_{h=1}^{H} \sum_{w=1}^{W} 
%     \left[ - y^n_c(h,w) \log (\hat{y}^n_c(h,w)) - (1-y^n_c )\log (1-\hat{y}^n_c(h,w)) \right]
% \end{equation}
where $\mathcal{L}_{pbce}$ is a pixel-wise binary cross entropy term and $\mathcal{L}_{dice}$ is a generalized dice loss \cite{dice_2017} commonly used to train segmentation models. 

The final loss term $\mathcal{L}$ is defined as a combination of the classification and segmentation loss terms summed over all $N$ cases in the training set such that:
\begin{equation}
    \mathcal{L} = \sum_{n=1}^N \mathcal{L}_{cls}(y_{l}^n, \hat{y}_{l}^n) + \mathbbm{1}_{y_{s}^n}*\mathcal{L}_{seg}(y_{s}^n, \hat{y}_{s}^n)
\end{equation}
where $\mathbbm{1}_{y_{s}^n} $ is an indicator variable such that $\mathbbm{1}_{y_{s}^n} = 1$ if pixel-wise labels exists for the $n$-th training case $X^n$.

\subsection{Annotation Granularity Definitions}
\label{sec:annotation_types}
We define four levels of annotation granularity: image-level labels extracted from text reports using NLP algorithms such as \cite{cheXpert}, image-level labels from radiologists, pixel-level annotations defined using bounding boxes around findings, and annotations drawn as polygons (contours). We note that we do not actually use models such as CheXPert \cite{cheXpert} to generate labels but manually control the amount of label noise that we introduce to the set of gold-standard annotations to mimic the NLP algorithm's behavior. Throughout this paper, we use the notation NLP@F1$_{X}$ to represent a model trained using a set of labels that are generated with an NLP algorithm with an F1 score of $X$. By controlling the F1 score of the NLP based label generator, we measure how the cheaper yet potentially inaccurate labels affect the CAD model performance and report the results in Section \ref{sec:classification}. 

Compared to image-level annotations, pixel-level annotations potentially provide more information to the model by localizing the lesions in X-ray images. We define different granularities in providing this localization information. Our intuition is that drawing coarse outlines such as bounding boxes around the lesion takes less effort and thus less time than drawing accurate contours around the lesions. To investigate this, we simulate bounding box annotations by generating them from our gold-standard contour labels. The generated bounding boxes provide pixel-level supervision to the model via $\mathcal{L}_{seg}$ where $y_{s}(h,w)=1$ for all pixels $(h,w)$ within the bounding box. 
%Figure \ref{fig:box_contour} compares the gold-standard contour labels to more cost efficient but coarse bounding box labels.

%\taesoo{will continue writing after we fix our segmentation method strategy}

% Our method consists of a model which handles classification and segmentation together, with loss formulation for each.
% Hence the model can be successfully trained regardless of which side of supervision is given.

% \Bparagraph{Model Architecture}
% ResNet-34~\cite{} as its backbone network.
% For the performance under usage of various backbone network, we refer the readers to the supplementary materials.

% The backbone network is followed by task-wise convolutional layers which produces localization heatmaps for each chest abnormalities, as~\cite{MDNet, }   \onwn{something from medical, with similar setup?}.

% We further compare the existing semantic segmentation methods~\cite{FCN, } trained on same experimental setting~\ref{}. 

%\Bparagraph{Loss Formulation}

%% file: experiments.tex
\subsection{Data and Settings}
The dataset used in this study contains a total of $121,356$ training samples where each image is a frontal view chest X-ray image with a set of gold-standard annotations provided by expert radiologists. Table \ref{tab:exp_db_stats} summarizes the number of positive and negative training/test samples for the ten chest abnormalities defined in this study. The test set consists of a total of $3,546$ independent cases also with gold-standard annotations.
For more details on the data and experimental settings used in this paper, please refer to the supplementary material. We report the average of three independent runs and the associated $95\%$ confidence intervals for all experiments reported in this paper.

\vspace*{-\baselineskip}
\input{tables/exp_db_stats}
\vspace*{-\baselineskip}

\subsection{Classification}
\label{sec:classification}
In this section, we present how the quantity, quality and granularity of annotation ultimately impacts the CAD model's classification performance.

\Bparagraph{Impact of Granularity: } The impact of annotation granularity on the resulting classification models is summarized in Table \ref{tab:cls_label_cost_size}. We use macro averaged AUCROC over ten lesions to measure the classification performance of the models. 

\input{tables/cls_label_cost_size}

Our experiments show that on relatively small scale datasets (less than 12K training samples in total), the models performed similarly even when additional supervision in the form of bounding boxes or contours were provided during training. However, when more training data points were available, additional pixel-wise supervision from bounding boxes or contours improves over the image-level models by over 2 points when using all 121K training cases. For providing pixel-level supervision to the classification model, we did not observe meaningful differences between models trained with bounding boxes and contours. Our results suggest that collecting bounding box level labels is a more cost effective alternative to collecting contours for classification problems.

Models trained with noisy labels consistently performed worse than the models supervised with ground truth provided by radiologists. However, the performance gap between the NLP@F1$_{0.80}$ and the image-level models was much smaller (less than $1$ point) when using 121K training samples. The performance gap between the two widens as less training data is available. The patterns are consistent with the literature on learning with noisy labels for medical image analysis \cite{KARIMI2020}. This suggests that using cheap labels that can be automatically extracted from text reports have the potential to yield accurate classification models given large enough training datasets.

\Bparagraph{Impact of Quality: } Given that the annotations extracted using NLP algorithms is potentially a very cost effective approach for training classification models for chest radiographs, we scrutinize the CAD model classification performance as the accuracy of NLP models improves. The impact of this label noise is summarized in Table \ref{tab:cls_label_cost_noise}.

\input{tables/cls_nlp_noise} 

As the automatic NLP based label extracting algorithms get more accurate, the resulting CAD classification models improve as well. Our results show that a relatively small gain of 0.05 in F1 score leads to considerable CAD model performance gains (92.9 of NLP@F1$_{0.80}$ vs 94.1 of NLP@F1$_{0.85}$). Using more accurate label generators improves data efficiency. Our results show that we can use only half the amount of data (60K) to achieve 92.2 AUCROC when using NLP@F1$_{0.85}$ whereas all 121K cases are needed to get to 92.9 AUCROC when using NLP@F1$_{0.80}$ algorithm. When using generated labels, it is important to invest in improving the NLP algorithm itself to maximize CAD model performance with the same amount of annotation budget.

\subsection{Segmentation}
\label{sec:segmentation}
In this section, we present our findings on how factors that determine annotation cost ultimately impact the segmentation performance of CAD models. We measure the segmentation performance using an aggregated Sørenson-Dice coefficient \cite{dice1945} macro averaged over all lesions. 
\input{tables/seg_label_cost_size}

\Bparagraph{Impact of Granularity and Quality: } In Table \ref{tab:seg_label_cost_size}, we report the segmentation performance difference between a model trained using pixel-level supervision extracted from bounding boxes and a model trained using gold-standard contours. The segmentation performances of both bounding-box and contour based models improve as more data is available. However, in contrast to the classification setting where the performance gap gradually decreased across different models as more data was available, the segmentation performance gap between the two models was kept consistent throughout all dataset scales. In fact, the results trend similarly with the experiments in Table \ref{tab:cls_label_cost_noise} where we measure performance across different noise levels. We observe that a bounding box is actually a polygon with much fewer degrees of freedom than a contour. We can view bounding boxes as lower quality (noisier) alternatives to contour annotations. The visualizations of model predictions in Figure 1 of the supplementary material highlight that the model trained with contours yields much cleaner segmentation results.

\begin{figure}[t]
\centering
     \includegraphics[width=0.70\textwidth]{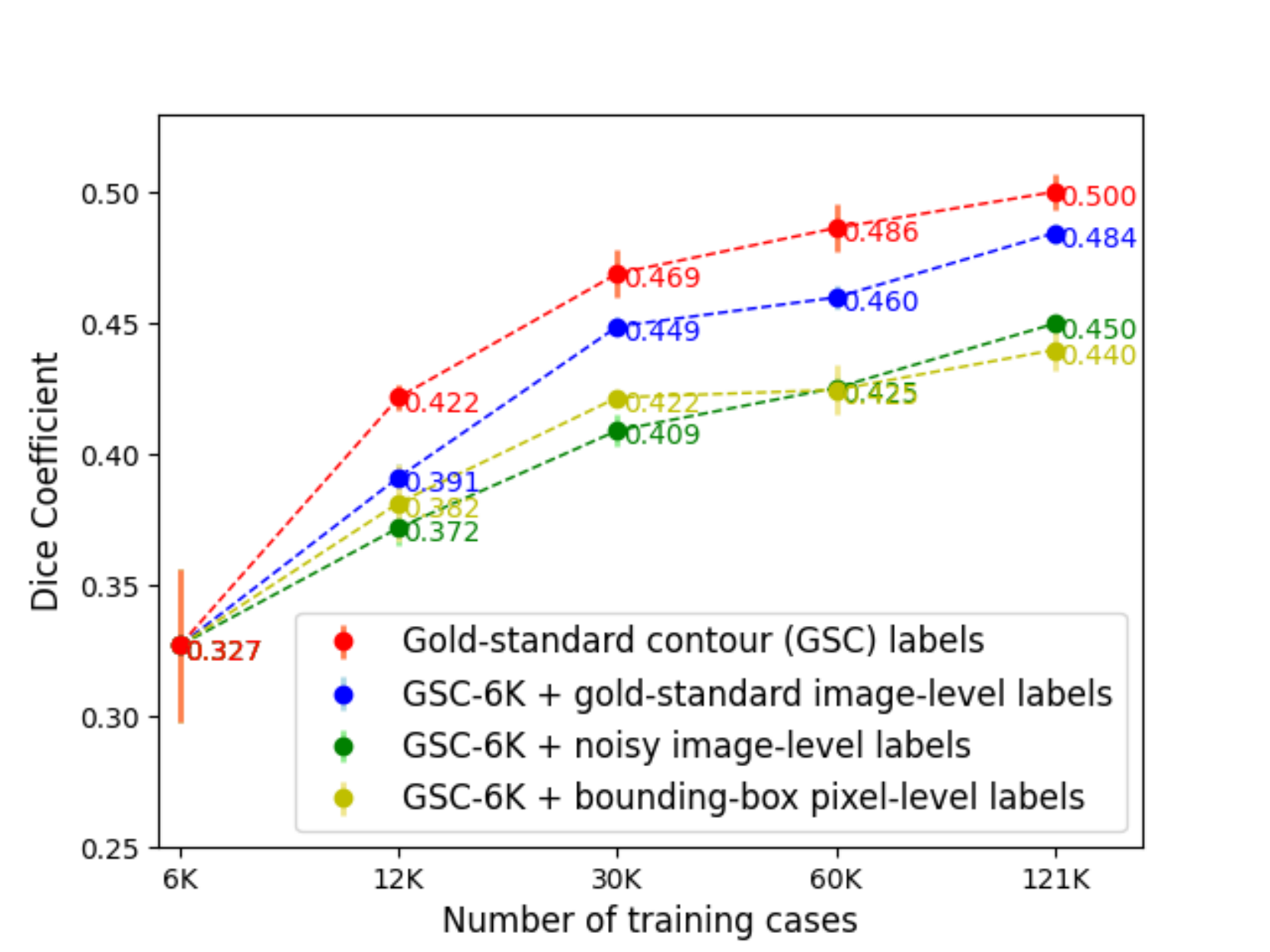}
    \caption{
    Effects of adding additional gold-standard contours, pixel-level information from bounding boxes, gold-standard image-level and noisy (NLP@F1$_{0.75}$) annotations on the segmentation model performance.
    %By combining image-level annotations with 6K images with contour labels, we can improve the model's segmentation performance.
    }
    \label{fig:adding_data}
\end{figure}

% \Bparagraph{Can We Mix Levels of Granularity and Save? } 
%\Bparagraph{High Performance from a Large Portion of Low-Cost Labels}
\Bparagraph{Improving Performance by Mixing Levels of Granularity:} 
We investigate whether we can train segmentation models in a more economic manner by adding large amounts of cost-efficient annotations to a small set of contour annotations. For example, as shown in Table \ref{tab:seg_label_cost_size}, we can improve Dice coefficient of the model trained with 6K training images by about $0.14$ points by adding $24$K extra training samples with contour annotations.
In our experiments, we observe that we can improve the same model (Contour-6K) by almost a similar amount ($0.12$) by adding the same $24$K extra training samples but with only \textit{image-level} annotations. In Figure \ref{fig:adding_data}, we show how adding various amounts of training images with only image-level annotations benefits the original segmentation model trained with small amounts (6K) of training data.
Interestingly, we observe that combining image-level annotations with contours is a better strategy than combining bounding boxes with contours.
At the largest dataset scale, adding bounding box annotations leads to a slightly worse segmentation model than adding the noisy NLP@F1$_{0.75}$ (Yellow in Figure \ref{fig:adding_data}) label.
% Figure~\ref{vis_comp_c} and \ref{vis_comp_d} shows that the bounding box annotations behave as label noise compared to gold-standard contours and lead a model produce over-segmented results.
% Comparing Figure~\ref{vis_comp_c} from Figure~\ref{vis_comp_a} shows that the high portion of image-level annotated data could guide a model to behave as successful as it is trained on data labeled in gold-standard contours.
For a more qualitative comparison of segmentation results of the models, please refer to Figure 1 from the supplementary material.
When there is not enough budget available for collecting large amounts of pixel-level annotations, the results suggest that collecting large amounts of image-level annotations and combining them with available pixel-level data is a cost efficient strategy for training segmentation models.

%\textbf{Qualitative results: Comparing heatmap output of model trained with contours/bbox/mix}

%% file: tables/exp_db_stats.tex
\begin{table*}[]
\centering
\setlength{\tabcolsep}{2.5pt}
\caption{Number of positive and negative samples for each chest abnormality used in this study. Atl: Atelectasis, Calc: Calcification, Cm: Cardiomegaly, Csn: Consolidation, Fib: Fibrosis, MW: Mediastinal widening, Ndl: Nodule, PEf: Pleural Effusion, Ppm: Pneumoperitoneum, Ptx: Pneumothorax}
\label{tab:exp_db_stats}
\begin{tabular}{c|cccccccccc}
Train & Atl    & Calc   & Cm     & Csn   & Fib    & MW     & Ndl    & PEf    & Ppm    & Ptx    \\
\toprule
Pos & 16939 & 9308 & 10602 & 43095 & 14704 & 1607 & 10500 & 21290 & 3108 & 6018 \\
Neg & 104417 & 112048 & 110754 & 78261 & 106652 & 119749 & 110856 & 100066 & 118248 & 115338 \\
\midrule
Test &   &  &  &  &  &  &  &  &  &  \\ \toprule  
Pos     & 272  & 267 & 187 & 664 & 272 & 42 & 694 & 510 & 152 & 437  \\
Neg     & 1842 & 1881 & 1880 & 1785 & 1793 & 2071 & 2351 & 2775 & 1961 & 3109 \\
\bottomrule
\end{tabular}
\end{table*}

%% file: tables/cls_label_cost_size.tex
\begin{table*}[t]
\centering
\setlength{\tabcolsep}{2.2pt}
%\begin{tabular}{r|cccccc}
\caption{
How annotation granularity impacts CAD model performance across various dataset scales.
Performance reported in AUCROC ($\times 100$) with 95\% confidence intervals.
}
\label{tab:cls_label_cost_size}
\begin{tabular}{c|cccccc}

 & \multicolumn{6}{c}{Dataset Size} \\
  & 1.2K (1\%) & 6K (5\%) & 12K (10\%) & 30K (25\%) & 60K (50\%) & 121K \\ \toprule
NLP @ $\text{F1}_{0.80}$ & 55.7 $\pm$ 5.0 & 66.9 $\pm$ 9.2 & 73.2 $\pm$ 3.0 & 84.5 $\pm$ 1.6 & 88.4 $\pm$ 1.8 & 92.9 $\pm$ 0.7  \\
Image Level & 65.3 $\pm$ 5.1 & 84.0 $\pm$ 3.2 & 90.4 $\pm$ 0.1  & 92.4 $\pm$ 0.2 & 92.9 $\pm$ 0.1  & 93.7 $\pm$ 0.7 \\
Bbox & 67.5 $\pm$ 1.9  & 81.2 $\pm$ 1.2 & 88.5 $\pm$ 0.7 & 94.3 $\pm$ 0.1  & 94.9 $\pm$ 0.3 & 95.8 $\pm$ 0.01 \\
Contours & 68.9 $\pm$ 1.3 & 82.1 $\pm$ 0.5 & 89.8 $\pm$ 1.3 & 94.6 $\pm$ 0.08  & 95.3 $\pm$ 0.03 & 95.9 $\pm$ 0.03 \\
\bottomrule
\end{tabular}

\end{table*}

%% file: tables/cls_nlp_noise.tex
\begin{table*}[t]
\centering
\setlength{\tabcolsep}{2.5pt}
%\begin{tabular}{r|cccccc}
\caption{How annotation noise commonly introduced by NLP derived labels impacts CAD model performance across various dataset scales. Performance reported in AUCROC with 95\% confidence intervals.}
\label{tab:cls_label_cost_noise}
\begin{tabular}{c|cccccc}
 & \multicolumn{6}{c}{Dataset Size} \\
 NLP F1 Score & 1.2K (1\%) & 6K (5\%) & 12K (10\%) & 30K (25\%) & 60K (50\%) & 121K \\ \toprule
0.75  & 52.9 $\pm$ 3.8 & 54.6 $\pm$ 2.9 & 71.6 $\pm$ 9.3 & 77.2 $\pm$ 4.2 & 86.2 $\pm$ 2.3 & 90.0 $\pm$ 0.1  \\
0.80 & 55.7 $\pm$ 5.0 & 66.9 $\pm$ 9.2 & 73.2 $\pm$ 3.0 & 84.5 $\pm$ 1.6 & 88.4 $\pm$ 1.8 & 92.9 $\pm$ 0.7  \\
0.85 & 57.7 $\pm$ 6.0 & 72.4 $\pm$ 3.1 & 83.2 $\pm$ 9.3 & 86.1 $\pm$ 1.1 &92.2 $\pm$ 0.5 & 94.1 $\pm$ 0.1 \\
0.90 & 61.8 $\pm$ 5.2 & 69.6 $\pm$ 7.7 & 82.9 $\pm$ 3.8 & 90.8 $\pm$ 2.8 & 93.2 $\pm$ 0.4 & 94.0 $\pm$ 0.3\\
0.95 & 60.1 $\pm$ 2.6 & 82.9 $\pm$ 0.6 & 87.5 $\pm$ 3.9 & 92.75 $\pm$ 0.3 & 93.3 $\pm$ 0.2 & 93.9 $\pm$ 0.8 \\
\bottomrule
\end{tabular}

\end{table*}

%% file: tables/seg_label_cost_size.tex
\begin{table*}[t]
\centering
\setlength{\tabcolsep}{2.5pt}
%\begin{tabular}{r|cccccc}
\caption{How annotation granularity impacts CAD model segmentation performance across various dataset scales. Performance measured in mean over per-lesion Dice Coefficients with 95\% confidence intervals.}
\label{tab:seg_label_cost_size}
\begin{tabular}{c|cccccc}

 & \multicolumn{6}{c}{Dataset Size} \\
  & 1.2K (1\%) & 6K (5\%) & 12K (10\%) & 30K (25\%) & 60K (50\%) & 121K \\ \toprule
Bbox & .046 $\pm$ .02 & .229 $\pm$ .02 & .341 $\pm$ .01 & .404 $\pm$ .01 & .421 $\pm$ .02 & .427 $\pm$ .02 \\
Contours & .111 $\pm$ .07 & .327 $\pm$ .03 & .422 $\pm$ .005 & .469 $\pm$ .009 & .486 $\pm$ .009 & .500 $\pm$ .007 \\
\bottomrule
\end{tabular}

\end{table*}

%% file: conclusion.tex
We presented a large-scale analysis of how annotation cost impacts CAD systems for classification and segmentation of chest abnormalities in frontal view chest X-rays.
We defined quantity, quality and granularity as the factors that determine annotation cost.
We found that the more cost efficient bounding boxes are as useful as the accurate contours when provided as additional supervision to the classification model.
We also showed that relatively small improvements to the label extracting algorithms lead to gains in classification performance.
Lastly, we reported that we can achieve strong segmentation performance by mixing image-level labels with only small amounts of pixel-level contour labels.
A limitation of the study is how we modeled the errors introduced by an NLP label extractor. We generated errors randomly at a given F1 operating point but the errors produced by the actual NLP model will be much more nuanced.
As part of future work, we wish to design a study with labels generated with actual NLP models.
As for the next follow-up study, we hope to analyze how the cost actually paid to the annotator changes depending on the method and amount of annotation.
Moreover, we hope to investigate whether our findings generalize to other modalities such as CT, breast mammography and MRIs.
We hope the results presented in this paper provide practical guidance for practitioners building automated CAD systems for chest radiographs to ultimately improve patient care by assisting healthcare providers and medical experts.

%% file: main.bbl
\begin{thebibliography}{10}
\providecommand{\url}[1]{\texttt{#1}}
\providecommand{\urlprefix}{URL }
\providecommand{\doi}[1]{https://doi.org/#1}

\bibitem{bustos2020padchest}
Bustos, A., Pertusa, A., Salinas, J.M., de~la Iglesia-Vay{\'a}, M.: Padchest: A
  large chest x-ray image dataset with multi-label annotated reports. Medical
  image analysis  \textbf{66},  101797 (2020)

\bibitem{dice1945}
Dice, L.R.: Measures of the amount of ecologic association between species.
  Ecology  \textbf{26}(3),  297--302 (July 1945)

\bibitem{He2016DeepRL}
He, K., Zhang, X., Ren, S., Sun, J.: Deep residual learning for image
  recognition. 2016 IEEE Conference on Computer Vision and Pattern Recognition
  (CVPR) pp. 770--778 (2016)

\bibitem{cheXpert}
Irvin, J., Rajpurkar, P., Ko, M., Yu, Y., Ciurea-Ilcus, S., Chute, C.,
  Marklund, H., Haghgoo, B., Ball, R., Shpanskaya, K., et~al.: Chexpert: A
  large chest radiograph dataset with uncertainty labels and expert comparison.
  In: Proceedings of the AAAI conference on artificial intelligence. vol.~33,
  pp. 590--597 (2019)

\bibitem{Jin2022}
Jin, K.N., Kim, E.Y., Kim, Y.J., Lee, G.P., Kim, H., Oh, S., Kim, Y.S., Han,
  J.H., Cho, Y.J.: Diagnostic effect of artificial intelligence solution for
  referable thoracic abnormalities on chest radiography: a multicenter
  respiratory outpatient diagnostic cohort study. European radiology
  \textbf{32}(5),  3469--3479 (2022)

\bibitem{KARIMI2020}
Karimi, D., Dou, H., Warfield, S.K., Gholipour, A.: Deep learning with noisy
  labels: Exploring techniques and remedies in medical image analysis. Medical
  Image Analysis  \textbf{65},  101759 (2020).
  \doi{https://doi.org/10.1016/j.media.2020.101759}

\bibitem{Li2020SelfLoopUA}
Li, Y., Chen, J., Xie, X., Ma, K., Zheng, Y.: Self-loop uncertainty: A novel
  pseudo-label for semi-supervised medical image segmentation. In: MICCAI
  (2020)

\bibitem{fcn_2015}
Long, J., Shelhamer, E., Darrell, T.: Fully convolutional networks for semantic
  segmentation. In: 2015 IEEE Conference on Computer Vision and Pattern
  Recognition (CVPR). pp. 3431--3440 (2015)

\bibitem{Rayner2019}
Oakden-Rayner, L.: Exploring large-scale public medical image datasets.
  Academic Radiology  \textbf{27} (11 2019). \doi{10.1016/j.acra.2019.10.006}

\bibitem{olatunji_2019}
Olatunji, T., Yao, L., Covington, B., Upton, A.: Caveats in generating medical
  imaging labels from radiology reports with natural language processing. In:
  International Conference on Medical Imaging with Deep Learning -- Extended
  Abstract Track. London, United Kingdom (08--10 Jul 2019)

\bibitem{Peng2021MedicalIS}
Peng, J., Wang, Y.: Medical image segmentation with limited supervision: A
  review of deep network models. IEEE Access  \textbf{9},  36827--36851 (2021)

\bibitem{CheXbert_emnlp2020}
Smit, A., Jain, S., Rajpurkar, P., Pareek, A., Ng, A., Lungren, M.: Combining
  automatic labelers and expert annotations for accurate radiology report
  labeling using {BERT}. In: Proceedings of the 2020 Conference on Empirical
  Methods in Natural Language Processing (EMNLP). pp. 1500--1519. Association
  for Computational Linguistics, Online (Nov 2020).
  \doi{10.18653/v1/2020.emnlp-main.117},
  \url{https://aclanthology.org/2020.emnlp-main.117}

\bibitem{dice_2017}
Sudre, C., Vercauteren, T., Ourselin, S., Cardoso, M.J.: Generalised dice
  overlap as a deep learning loss function for highly unbalanced segmentations.
  vol.~2017, pp. 240--248 (09 2017). \doi{10.1007/978-3-319-67558-9_28}

\bibitem{wang_cxr14}
Wang, X., Peng, Y., Lu, L., Lu, Z., Bagheri, M., Summers, R.M.: Chestx-ray8:
  Hospital-scale chest x-ray database and benchmarks on weakly-supervised
  classification and localization of common thorax diseases. In: CVPR. pp.
  3462--3471. IEEE Computer Society (2017)

\bibitem{CALLI2021}
Çallı, E., Sogancioglu, E., {van Ginneken}, B., {van Leeuwen}, K.G., Murphy,
  K.: Deep learning for chest x-ray analysis: A survey. Medical Image Analysis
  \textbf{72},  102125 (2021).
  \doi{https://doi.org/10.1016/j.media.2021.102125},
  \url{https://www.sciencedirect.com/science/article/pii/S1361841521001717}

\end{thebibliography}
